%% file: main.tex
\documentclass{article}

\PassOptionsToPackage{numbers, compress}{natbib}
\usepackage[preprint]{neurips_2019}

\usepackage[utf8]{inputenc} % allow utf-8 input
\usepackage[T1]{fontenc}    % use 8-bit T1 fonts
\usepackage{hyperref}       % hyperlinks
\usepackage{url}            % simple URL typesetting
\usepackage{booktabs}       % professional-quality tables
\usepackage{amsfonts}       % blackboard math symbols
\usepackage{nicefrac}       % compact symbols for 1/2, etc.
\usepackage{microtype}      % microtypography
\usepackage{subcaption}
\usepackage[pdftex]{graphicx}
\usepackage{colortbl}
\usepackage{color}

\newcommand*\samethanks[1][\value{footnote}]{\footnotemark[#1]}

\usepackage{amsmath}
\usepackage{algorithm}
\usepackage[noend]{algorithmic}

\usepackage{caption}
\usepackage[noend,algo2e]{algorithm2e}
\newcommand \from \leftarrow
\usepackage{listings}
\usepackage{xcolor}
\usepackage{textcomp}
 
\definecolor{codegreen}{rgb}{0,0.6,0}
\definecolor{codegray}{rgb}{0.5,0.5,0.5}
\definecolor{codepurple}{rgb}{0.58,0,0.82}
 
\lstdefinestyle{python}{
    language=Python,
    columns=fullflexible,
    commentstyle=\color{codegreen},
    keywordstyle=\color{blue},
    numberstyle=\tiny\color{codegray},
    stringstyle=\color{codepurple},
    basicstyle=\ttfamily\scriptsize,
    breakatwhitespace=false,         
    breaklines=true,                 
    captionpos=b,                    
    keepspaces=true,                 
    numbers=left,                    
    numbersep=5pt,                  
    showspaces=false,                
    showstringspaces=false,
    showtabs=false,                  
    tabsize=2,
    upquote=true,
}
 
\usepackage{array}
\newcolumntype{L}[1]{>{\raggedright\let\newline\\\arraybackslash\hspace{0pt}}p{#1}}
\newcolumntype{C}[1]{>{\centering\let\newline\\\arraybackslash\hspace{0pt}}p{#1}}
\newcolumntype{R}[1]{>{\raggedleft\let\newline\\\arraybackslash\hspace{0pt}}m{#1}}

\title{Composable Effects for Flexible and Accelerated Probabilistic Programming in NumPyro}

\author{%
  Du Phan \thanks{Equal contribution} \\
  \texttt{phandu@postech.ac.kr} \\
  \And
  Neeraj Pradhan \samethanks \\
  \texttt{npradhan@uber.com} \\
  Uber AI \\
  \And
  Martin Jankowiak \\
  \texttt{jankowiak@uber.com} \\
}

\begin{document}

\maketitle

\begin{abstract}
NumPyro is a lightweight library that provides an alternate NumPy backend to the Pyro probabilistic programming language with the same modeling interface, language primitives and effect handling abstractions. Effect handlers allow Pyro's modeling API to be extended to NumPyro despite its being built atop a fundamentally different JAX-based functional backend. 
%JAX provides a high-level tracing mechanism that enables JIT compilation for hardware acceleration, automatic differentiation, and vectorization. 
%In this work, we demonstrate how composing these effects allows us to implement powerful inference algorithms and utilities that can be scaled using SIMD hardware. 
In this work, we demonstrate the power of composing Pyro's effect handlers with the program transformations that enable hardware acceleration, automatic differentiation, and vectorization in JAX. 
In particular, NumPyro provides an iterative formulation of the No-U-Turn Sampler (NUTS) that can be end-to-end JIT compiled, yielding an implementation that is much faster than existing alternatives in both the small and large dataset regimes. 
\end{abstract}

%NumPyro is a lightweight library for probabilistic programming that provides an alternate NumPy backend for the Pyro probabilistic programming language. NumPyro provides the same modeling interface, language primitives and effect handling abstractions as Pyro. Effect handlers allow Pyro's modeling API to be extended to NumPyro despite its being built atop a fundamentally different JAX-based functional backend. JAX provides a high-level tracing mechanism implemented as an effectful operation over NumPy primitives to enable JIT compilation for hardware acceleration, automatic differentiation, and vectorization. In this work, we demonstrate how composing these effects allows us to implement powerful inference algorithms and utilities that can be scaled using SIMD hardware. In particular, NumPyro provides an  iterative formulation of the No-U-Turn Sampler (NUTS) that can be end-to-end JIT compiled, yielding an implementation that is much faster than existing alternatives in both the big and small dataset regimes. 

\section{Introduction}

Many probabilistic programming languages (PPLs) \cite{van2018introduction} are embedded as a DSL within a host language. The advantage of embedding is availability of host language infrastructure along with frameworks for automatic differentiation and hardware acceleration. Within the Python community, some examples of embedded PPLs include Pyro \cite{bingham2019} and ProbTorch \cite{narayanaswamy2017learning} based on PyTorch \cite{paszke2017automatic}, TensorFlow Probability \cite{dillon2017tensorflow} and Edward2 \cite{tran2018simple} based on TensorFlow, and PyMC3 \cite{salvatier2016} based on Theano. NumPyro is a package for probabilistic programming built atop JAX \cite{jax2018github, frostig2018}, which is a high-level tracing library for program transformations (e.g.~automatic differentiation, vectorization and JIT compilation) of Python and NumPy functions. Thus NumPyro enables users to write probabilistic programs using familiar NumPy arrays and operations. 

NumPyro is built around the same effect handling abstraction as Pyro. Effect handlers provide a way to inject effectful computation into primitive statements in a probabilistic program, e.g. recording the random choices made in an execution trace. In NumPyro these effects can be easily composed with the JAX tracer that operates at the level of NumPy operations and its own set of primitives for control flow. This allows us to expose a modeling language that is the same as in Pyro. Under the hood, inference algorithms can use effect handlers to inspect and modify program behavior and freely compose with JAX transformations to speed up critical subroutines via parallelization and JIT compilation. As an example (see Sec.~\ref{sec:nuts}), we implement an iterative version of the No-U-Turn Sampler that can leverage JAX's \verb|jit| transformation for end-to-end compilation and optimization by the XLA compiler \cite{xla}.

\section{Support for Pyro's Modeling Interface}
%Different frameworks have their own particular strengths and limitations and may affect the design of the embedded DSL. Despite some core differences in the underlying backend, we have 
NumPyro retains the same language primitives and modeling and inference interface as in Pyro.\footnote{A generic API for modeling and inference for dispatch to different Pyro backends can be found at \url{https://github.com/pyro-ppl/pyro-api}} In particular, NumPyro supports \verb|sample| and \verb|param| statements that allow users to designate random variables and learnable parameters, respectively. It also has effect handlers like \verb|trace|, \verb|replay| and \verb|condition| to provide nonstandard interpretations to these statements. Table~\ref{tab:handlers} lists some commonly used effect handlers. 

Effect handlers have emerged as a composable abstraction for program transformations in PPLs \cite{bingham2019, moore2018, plotkin}. We have found the effect handling abstraction to be particularly useful in designing a common interface to probabilistic programming, despite fundamental differences in the underlying backend. While PyTorch tries to accommodate much of Python's dynamism and facilities for object-oriented programming with mutable objects (e.g.~PyTorch optimizers update parameters in-place), JAX encourages a functional style of programming as required by the tracer. As an example, unlike PyTorch, JAX uses a functional pseudo-random number generator \cite{johnson2019prng}, which mandates passing an explicit random number generator key (\verb|PRNGKey|) to distribution samplers. In practice, NumPyro inference algorithms take a single \verb|PRNGKey| from the user that is split to generate new keys when passed to downstream functions. This does not result, however, in any change to probabilistic programs formulated in NumPyro, as this splitting mechanism is abstracted into a \verb|seed| handler that operates on \verb|sample| statements (see Table~\ref{fig:logreg}).

\begin{table}[t!]
    \centering
    \begin{tabular}[t!]{|R{17mm}|L{36mm}|L{33mm}|L{33mm}|}
\hline
\cellcolor[gray]{0.85} & 
\multicolumn{1}{c|}{\textbf{seed} \cellcolor[gray]{0.95}} & 
\multicolumn{1}{c|}{\textbf{trace} \cellcolor[gray]{0.95}} & 
\multicolumn{1}{c|}{\textbf{condition} \cellcolor[gray]{0.95}}\\
\hline
\small Primitives Affected & \small {\texttt{sample}} & \small \texttt{sample}, \texttt{param} & \small \texttt{sample} \\
\hline
\small Description & 
\small Seeds \texttt{fn} with a \texttt{PRNGKey}. Every call to \texttt{sample} inside \texttt{fn} results in splitting of \texttt{PRNGKey} to generate a fresh seed for subsequent calls. &

\small Records the input, output, and function calls inside of \texttt{sample}, \texttt{param} statements in \texttt{fn}. &
\small Conditions unobserved \texttt{sample} sites in \texttt{fn} to values in \texttt{param}. \\
\hline 
\small Usage &
\scriptsize \texttt{seed(fn, rng\_key)(...)} & 
\scriptsize \texttt{trace(fn).get\_trace(...)} & 
\scriptsize \texttt{condition(fn, param)(...)} \\
\hline
\end{tabular}
\vspace{0.8em}
\caption{Examples of effect handlers: the primitive statements affected by each handler, the effect added, and usage w.r.t.~the original function \texttt{fn} with its arguments denoted by ellipsis.}
\label{tab:handlers}
\end{table}

\input{transformations}

\input{experiment}
\input{discussion}

\section{Acknowledgments}

We would like to thank Noah Goodman for feedback, and the JAX development team---in particular Matthew Johnson and Peter Hawkins---for their invaluable help with many JAX issues and feature requests.

\renewcommand{\bibsection}{\section{\refname}}
\medskip
\small
\bibliographystyle{unsrtnat}
\bibliography{numpyro}

\newpage
\appendix
\input{appendix}
\end{document}

%% file: transformations.tex
\section{Leveraging JAX Transformations in Inference Subroutines}
\label{transformations}

JAX \cite{frostig2018} is a Python library that provides a high-level tracer for implementing transformations of programs in Python and NumPy. Currently, three main transformations are available: i) automatic differentiation (\verb|grad|); ii) JIT compilation (\verb|jit|) to multiple backends using XLA; and iii) automatic vectorization (\verb|vmap|). \citet{frostig2018} note that ML workloads are composed of many pure-and-statically-composed (PSC) subroutines that are good candidates for acceleration. This is also true of the inference subroutines that lie at the core of NumPyro. While NumPyro's frontend---i.e.~its modeling and inference API---is close to Pyro, we have taken care to ensure that the core inference algorithms and utilities are purely functional so that we can make extensive use of JAX transformations like \verb|jit| and \verb|vmap|. This allows us to implement highly optimized and parallelizable subroutines. We provide two such examples: i) composing \verb|jit| and \verb|grad| to implement an iterative version of the NUTS sampler that can be end-to-end JIT compiled by JAX; and ii) composing \verb|vmap| with effect handlers like \verb|trace| or \verb|condition| to implement vectorized subroutines.

%Composing JAX transformations allows us to implement highly optimized inference subroutines. For instance, being able to compose \verb|jit| and \verb|grad| allows us to implement a JIT-compiled iterative version of the No-U-Turn Sampler (NUTS) (see Sec~\ref{sec:nuts}). JAX's \verb|vmap| allows us to vectorize computations in various inference utilities, e.g.~for predictions and log likelihood computations.

\input{nuts_rev2}

\subsection{Vectorizing Subroutines with \texttt{vmap}}

Many utilities and subroutines for inference, e.g.~model prediction, Monte Carlo estimation, or running MCMC chains, can be batched to make use of SIMD vectorization. In many frameworks this kind of batching requires laborious manual threading and/or significant cognitive overhead in managing explicit batch dimensions. JAX provides a vectorizing map (\verb|vmap|) transformation that makes it easy to represent batched computations as mapping over function arguments along an outermost axis. This requires no changes to the underlying code but maintains the efficiency of manual batching.

Since JAX transformations are fully composable with Pyro's effect handlers like \verb|seed|, \verb|trace|, and \verb|condition|, and since the latter are implemented within the Python runtime and thus traceable, \verb|vmap| becomes very powerful.
As an example, Fig.~\ref{fig:logreg} shows how we can use \verb|vmap| to batch three common computations: i) sampling from the prior; ii) sampling from the posterior predictive distribution; iii) and computing log-likelihoods. Note that without \verb|vmap| we would need to explicitly handle an additional batch dimension within the \verb|logistic_regression| model and the utility functions in Fig.~\ref{fig:pred}, which is particularly cumbersome for more involved models.  As a final example, in Stochastic Variational Inference (SVI) \cite{hoffman2013stochastic}, we optimize a loss function that is a Monte Carlo estimate of the Evidence Lower Bound (ELBO). This requires running the model as well as the inference network multiple times, all of which can be elegantly parallelized using \verb|vmap| (see Appendix~\ref{sec:vectorized-elbo}).

\begin{figure*}
\begin{subfigure}[t]{0.43\textwidth}\centering
\begin{lstlisting}
def logistic_regression(x, y=None):
  ndims = np.shape(x)[-1]
  m = sample('m', Normal(0., np.ones(ndims)))
  b = sample('b', Normal(0., 1.))
  return sample('y', Bernoulli(logits=x @ m + b), obs=y)
\end{lstlisting}
\vspace{2em}
\caption{A model for logistic regression in NumPyro}
\label{fig:log-reg}
\end{subfigure}
~  
~
\begin{subfigure}[t]{0.56\textwidth}\centering
\begin{lstlisting}
def predict_fn(rng_key, param, *args):
  conditioned_model = condition(logistic_regression, param)
  return seed(conditioned_model, rng_key)(*args)

def ll_fn(rng_key, param, *args):
  f_traced = trace(predict_fn)
  obs = f_traced.get_trace(rng_key, param, *args)['y']
  return np.sum(obs_node['fn'].log_prob(obs['value']))

\end{lstlisting}
\caption{Utilities for prediction and log-likelihood computation using \texttt{seed}, \texttt{trace} and \texttt{condition} handlers}
\label{fig:pred}
\end{subfigure}

\begin{subfigure}{\textwidth}
\begin{lstlisting}
# x -> input dataset
# samples -> a dict of samples from the posterior distribution
# rng_keys.. -> batch of PRNGKeys

prior_predictive = vmap(lambda rng_key: seed(logistic_regression, rng_key)(x))(rng_keys_sim)
posterior_predictive = vmap(lambda rng_key, param: predict_fn(rng_key, param, x))(rng_keys_pred, samples)
log_likelihood = vmap(lambda rng_key, param: ll_fn(rng_key, param, x, y))(rng_keys_pred, samples)
exp_log_likelihood = logsumexp(log_likelihood) - np.log(num_samples)
\end{lstlisting}
\caption{Vectorized sampling from the prior and posterior predictive with log-likelihood computation using \texttt{vmap}. A complete code listing (including inference) is provided in Appendix~\ref{app:logreg-listing}.}
\label{vec-pred}
\end{subfigure}
\caption{A simple logistic regression model. The modeling language is the same as in Pyro.}
\label{fig:logreg}
\end{figure*}

%\section{JIT Compilation of Inference Subroutines}

%\citet{frostig2018} mention that ML workloads are composed of many pure-and-statically-composed (PSC) sub-routines which can be JIT compiled. This is also true of inference subroutines that lie at the core of NumPyro. While NumPyro's frontend---i.e.~its modeling and inference API---are close to Pyro, we have taken care to ensure that the core inference algorithms and utilities are purely functional so as to extensively leverage JAX transformations like `jit` and `vmap`. We provide two such examples -- an iterative version of the NUTS sampler which can be end-to-end JIT compiled by JAX, and using `vmap` to 

%% file: nuts_rev2.tex
\subsection{Iterative No-U-Turn Sampler (NUTS)}
\label{sec:nuts}
	The No-U-Turn Sampler (NUTS) \cite{JMLR:v15:hoffman14a} is an extension of the Hamiltonian Monte Carlo (HMC) algorithm \cite{duane1987hybrid,nealhmc}, which can efficiently sample from high-dimensional continuous probability distributions. NUTS adaptively sets the trajectory length parameter in HMC, which along with the adaptation of the step size and mass matrix parameters, ensures that HMC runs efficiently on a variety of models without extensive hand-tuning. This provides a highly attractive black-box inference algorithm for PPLs to have in their toolkit.

%A key component of this algorithm is the recursive \verb|BuildTree| procedure, a simplified version of which is presented in Algorithm~\ref{alg:recursive-nuts} in the appendix. To build a tree at depth $d$, it builds two subtrees at depth $d-1$ and combines them. This recursive procedure also takes care to ensure that the memory consumption scales as $\mathcal{O}(\log{}N)$ (where $N = 2^d$) rather than $\mathcal{O}(N)$ by only storing $\mathcal{O}(1)$ information per subtree. This is important because storing all $n$ momentum-position pairs might be prohibitive for large models. 

To JIT compile NUTS sampling, we need to \verb|jit| transform a key component of the algorithm, namely the \verb|BuildTree| subroutine that recursively builds an implicit balanced binary tree by running the \verb|LeapFrog| integrator (Appendix~\ref{iterative-nuts-algo}). While this can be written as a PSC subroutine, tracing it is hard for two reasons. First, the form of the \verb|LeapFrog| integrator requires us to JIT through a gradient computation. JAX can handle this, since transformations like \verb|jit| and \verb|grad| are composable.\footnote{Note, however, that for example PyTorch's tracing JIT does not allow for this.} Second---and more problematically---the complex control flow of the recursive formulation cannot be traced for JIT compilation in JAX.\footnote{This obstacle was also noted in \citet{tran2018simple}.} 

An alternative would be to JIT compile a single \verb|LeapFrog| step or the potential energy function.\footnote{This is the approach adopted by the NUTS implementation in Pyro.} However, drawing a single sample involves many \verb|LeapFrog| steps, and the overhead in terms of Python function dispatch calls is significant. In addition, this approach significantly reduces opportunities for operator fusion in XLA compilation. To overcome these limitations, we propose an \emph{iterative} version of the NUTS algorithm that can be fully JIT compiled. In particular, this involves converting the \verb|BuildTree| procedure into an iterative procedure, paving the way for a NUTS implementation that can take full advantage of XLA acceleration. As we demonstrate in benchmarking experiments in Sec.~\ref{experiment}, the result is an algorithm that is much faster than existing implementations. More details on the algorithm are available in Appendix~\ref{iterative-nuts-algo}.

%% file: experiment.tex
\section{Experiments}
\label{experiment}

We compare the performance of NumPyro's NUTS implementation with that of other frameworks (Stan \cite{carpenter2017stan, allen_riddell_2018_1456206} and Pyro) in both the small and large data regimes. Recall that NumPyro's NUTS implementation is end-to-end JIT compiled, while in Pyro only the potential energy computation is compiled. We use three benchmark models: i) a Hidden Markov Model (HMM) on a small synthetic dataset; ii) logistic regression on the \verb|Forest CoverType| dataset \cite{Dua:2019}; and iii) a sparse kernel interaction model (SKIM) \cite{pmlr-v97-agrawal19a} on synthetic datasets with varying dimensionalities. Refer to Appendix \ref{sec:experimental-detail} for details on the benchmarking experiments.

%Taking into account the fact that the number of leapfrog steps to draw a sample varies during sampling, the measure which we will use is time per leapfrog step.

\begin{figure}
\centering
\begin{subfigure}[t!]{0.45\textwidth}
%\hskip-3.0cm
%\begin{subtable}[!h]{0.5\textwidth}
	\begin{tabular*}{\textwidth}{@{\extracolsep{\fill}} lrr}
		\toprule
	\small 	Framework      &  \small   HMM\footnotemark & \small  COVTYPE  \\
		\midrule
		\small Stan (64-bit CPU)     & \small   0.53 &  \small  135.94  \\
		%\small Edward2 (32-bit CPU)  & \small     - &   \small  50.39  \\
		%\small Edward2 (GPU)  & \small      - &  \small    8.28  \\
		\small Pyro (32-bit CPU)     & \small  30.51 &   \small  32.76  \\
		\small Pyro (GPU)     & \small - &    \small  3.36  \\
		\midrule
		\small NumPyro (32-bit CPU)  & \small \textbf{0.09} & \small 30.11  \\
		\small NumPyro (64-bit CPU)  & \small 0.15 & \small 71.18  \\
		\small NumPyro (GPU)  & \small - & \small \textbf{1.46}  \\
		\bottomrule
		\vspace{1.5em}
	\end{tabular*}
\caption{Time (ms) per leapfrog step in different frameworks.}
\label{table:experiment}

%\caption{Time per leapfrog step in different frameworks (in milliseconds).}\label{table:experiment}
%\end{subtable}

%\label{table:experiment}
\end{subfigure}
\quad
\begin{subfigure}[t!]{0.45\textwidth}\centering
	\centering
	\includegraphics[width=0.90\textwidth]{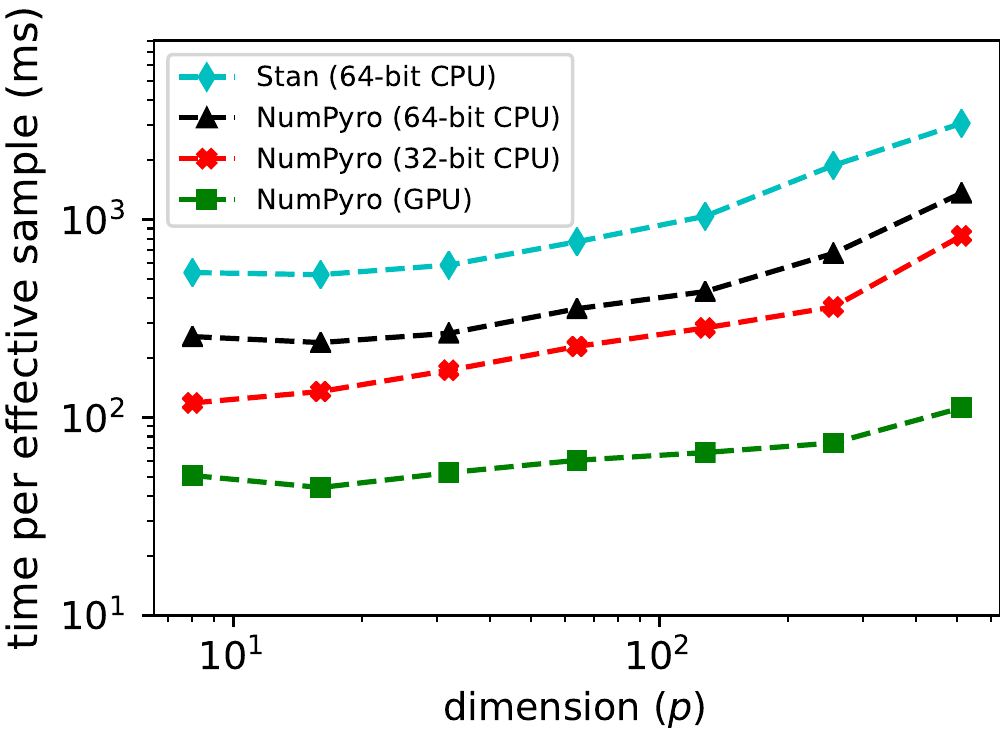}
	\caption{Time (ms) per effective sample for SKIM as the dimensionality of the dataset ($p$) is varied.}
	\label{fig:sparse_reg}
\end{subfigure}
\caption{Empirical evaluation of time taken by NumPyro's Iterative NUTS with respect to other frameworks.}

\end{figure}

\paragraph{Hidden Markov Model} Since we use a small dataset for this experiment, we expect poor performance on the GPU; consequently we limit ourselves to a CPU-only comparison. Note that, although the dataset is small, the potential energy computation involves a loop that can be expensive to differentiate through. From Table \ref{table:experiment}, we see that for the HMM, NumPyro is around $340$X faster than Pyro and $6$X faster than Stan. The iterative procedure in Algorithm \ref{alg:iterative-nuts} introduces 
%additional \verb|BitCount| operators that add 
insignificant overhead, and the end-to-end compilation allows XLA to output highly optimized code. 

\paragraph{Logistic Regression} For this dataset, which contains more than half a million datapoints, GPU acceleration significantly outperforms the CPU, as expected. The time spent in computing gradients in the \verb|LeapFrog| integrator exceeds the time spent building the tree or computing the terminating condition. Since the bottleneck primarily lies in large tensor operations, we expect the difference between the various GPU implementations to be narrower. Nevertheless, on this problem NumPyro is about 2X faster than Pyro.

\footnotetext{
	By conducting $5$ runs with different random seeds, $1000$ warmup steps and $1000$ samples for each run, the average effective sample size for Stan, NumPyro $32$-bit, and NumPyro $64$-bit are $652$, $556$, and $788$ respectively.
	}

\paragraph{Sparse Kernel Interaction Model} SKIM is Bayesian model for sparse regression that can be used to discover pairwise interactions in high dimensional data. Since the sparsity-inducing prior introduces a latent variable for each of the $p$ input dimensions, this represents a difficult inference problem when
$p$ is large. Fig.~\ref{fig:sparse_reg} shows how the time per effective sample using NUTS scales with the dimensionality of the dataset for Stan and NumPyro. We observe that NumPyro has consistently lower overhead as compared to Stan. NumPyro offers the flexibility to run inference in single or double precision as well as on different backends such as CPU, GPU, or TPU. While inference with double precision yields a higher effective sample size on average, it is not enough to compensate for the higher time taken to run inference, and hence the time per effective sample is lower for single precision. This model also particularly benefits from GPU acceleration, resulting in a major improvement in execution time when compared to the CPU backend.

%% file: discussion.tex
\section{Summary}

We describe NumPyro, a package for probabilistic programming using Python and NumPy that uses JAX transformations under the hood for hardware acceleration, automatic differentiation, and vectorization. NumPyro has a functional core, where inference subroutines are pure-and-statically composed functions that can be traced by JAX for parallelization and JIT compilation. These subroutines also make use of effect handlers to inspect and transform probabilistic programs. Effect handlers operate on core language primitives within the Python runtime, are transparent to the JAX tracer, and are, therefore, fully composable with JAX's transformations. This composability allows us to offer the same modeling language as Pyro, and at the same time leverage JAX tranformations to parallelize and JIT compile inference subroutines for significant speed ups. In particular we show that the judicious application of these program transformations allows us to implement  
an iterative version of the NUTS algorithm that offers strong performance on both the CPU (for small models) and the GPU (for larger models).

%As examples, we show that this allows us to implement an iterative version of the NUTS sampler that offers state of the art performance on both the CPU (for small sized problems) and GPU, and automatically batch various inference utilities, e.g. doing batched sampling from the prior and the posterior predictive, and vectorized computation of the Evidence Lower Bound (ELBo) estimate in SVI with no additional changes to the functions being vectorized.

%% file: appendix.tex
%\section{Appendix}
\input{iterative_nuts_algo.tex}

\section{Code for Vectorized Sampling - Logistic Regression}
\label{app:logreg-listing}

\begin{minipage}{\linewidth}
\begin{lstlisting}[caption={Using \texttt{vmap} to vectorize three common inference subroutines: i) sampling from the prior; ii) sampling from the posterior predictive distribution; and iii) computing log-likelihoods.},captionpos=b]
from jax import random, vmap
import jax.numpy as np
from jax.scipy.special import logsumexp

import numpyro
import numpyro.distributions as dist
from numpyro.handlers import condition, seed, trace
from numpyro.infer import MCMC, NUTS

def logistic_regression(x, y=None):
    ndims = np.shape(x)[-1]
    m = numpyro.sample('m', dist.Normal(0., np.ones(ndims)))
    b = numpyro.sample('b', dist.Normal(0., 1.))
    return numpyro.sample('y', dist.Bernoulli(logits=x @ m + b), obs=y)

def predict_fn(rng_key, param, *args):
    conditioned_model = condition(logistic_regression, param)
    return seed(conditioned_model, rng_key)(*args)

def loglik_fn(rng_key, params, *args):
    tr = trace(predict_fn).get_trace(rng_key, params, *args)
    obs_node = tr['y']
    return np.sum(obs_node['fn'].log_prob(obs_node['value']))

# Generate random data
true_coefs = np.array([1., 2., 3.])
x = random.normal(random.PRNGKey(0), (100, 3))
dim = 3
y = dist.Bernoulli(logits=x @ true_coefs).sample(random.PRNGKey(3))

# Run inference to generate samples from the posterior
num_warmup, num_samples = 500, 500
kernel = NUTS(model=logistic_regression)
mcmc = MCMC(kernel, num_warmup, num_samples)
mcmc.run(random.PRNGKey(1), x, y=y)
samples = mcmc.get_samples()

# Generate batch of PRNGKeys
rngs_sim = random.split(random.PRNGKey(2), num_samples)
rngs_pred = random.split(random.PRNGKey(3), num_samples)

# Prediction and log likelihood
prior_predictive = vmap(lambda rng_key: seed(logistic_regression, rng_key)(x))(rng_keys_sim)
posterior_predictive = vmap(lambda rng_key, param: predict_fn(rng_key, param, x))(rng_keys_pred, samples)
log_likelihood = vmap(lambda rng_key, param: loglik_fn(rng_key, param, x, y))(rng_keys_pred, samples)
expected_log_likelihood = logsumexp(log_likelihood) - np.log(num_samples)
\end{lstlisting}
\end{minipage}

\input{experiment_detail.tex}

\input{vectorized_elbo}

%% file: iterative_nuts_algo.tex
\section{Iterative NUTS - Algorithm Details}
\label{iterative-nuts-algo}

Computing a trajectory in NUTS involves a doubling procedure where at each iteration we run the \verb|LeapFrog| integrator for twice the number of steps taken in the previous iteration, with the direction (forward or reverse) chosen randomly. This has the effect of building an implicit balanced binary tree. The doubling process is terminated when a subtrajectory from the leftmost to the rightmost node of any balanced subtree begins to double back on itself. 

Existing NUTS implementations use a recursive tree building formulation, the \verb|BuildTree| subroutine, to double the trajectory length (see \citet[Algorithm 6]{JMLR:v15:hoffman14a}). A simplified version of this subroutine is presented in Algorithm~\ref{alg:recursive-nuts}. To build a tree at depth $d$, it builds two subtrees at depth $d-1$ and combines them. This recursive procedure also takes care to ensure that memory usage scales as $\mathcal{O}(\log{}N)$ (where $N = 2^d$) rather than $\mathcal{O}(N)$ by only storing $\mathcal{O}(1)$ data per subtree. This is important because storing all $N$ momentum-position pairs might be prohibitive for large models. The key to JIT compiling NUTS sampling is the ability to convert the recursive \verb|BuildTree| subroutine into an iterative implementation.

The iterative version of \verb|BuildTree| takes an initial node (position-momentum pair) $z_{-1}$ and a tree depth $d$ argument, and runs the \verb|LeapFrog| integrator for $N=2^d$ steps. We will be using $0$-based indexing in the following discussion. 

\paragraph{Checking the U-Turn Condition} For node $z_n$, we need to check the U-Turn condition with respect to the leftmost nodes of any binary subtree for which $z_n$ is the rightmost node. Let us denote the indices for these candidate nodes by $\mathcal{C}(n)$, and the binary representation of $n$ by $b(n)$. Indices in $\mathcal{C}(n)$ have the same binary representation as $b(n)$ except that trailing contiguous $1$s in $b(n)$ are progressively masked by $0$; e.g. for $n=11$, $b(11) = (1011)_2$, and the set of candidate nodes for checking the U-Turn condition are indexed by $\mathcal{C}(11) = \{(1010)_2, (1000)_2\} = \{8, 10\}$. Note that this implies that we only need to check the U-Turn condition at odd-numbered nodes against a subset of previous even-numbered nodes.

This allows us to iteratively build the binary tree by running the \verb|LeapFrog| integrator for $2^d$ steps, and terminating early if the U-Turn condition stated above is satisfied. However, in a naive implementation we would still need $\mathcal{O}(N)$ memory, since we would need to store the position-momentum pairs at each step of the integrator, which would be an unacceptable regression from the $\mathcal{O}(\log{}N)$ memory requirement of the recursive algorithm. 

\paragraph{Memory Efficiency} To tackle the issue of memory efficiency, we will use an array $S$ to store only even numbered nodes $z_k$ at index $i=\mathrm{\texttt{BitCount}}(k)$. As we iteratively build the tree, nodes in $S$ may be overwritten so that at step $n$, index $i$ stores the largest even node $z_{k}$ such that $k<n$ and  $\verb|BitCount|(k)=i$. Note that the maximum size of the array $S$ is $d$ because the largest bit count of even numbers less than $2^d$ is $d - 1$.

\begin{figure}
	\centering
	\includegraphics[width=0.90\textwidth]{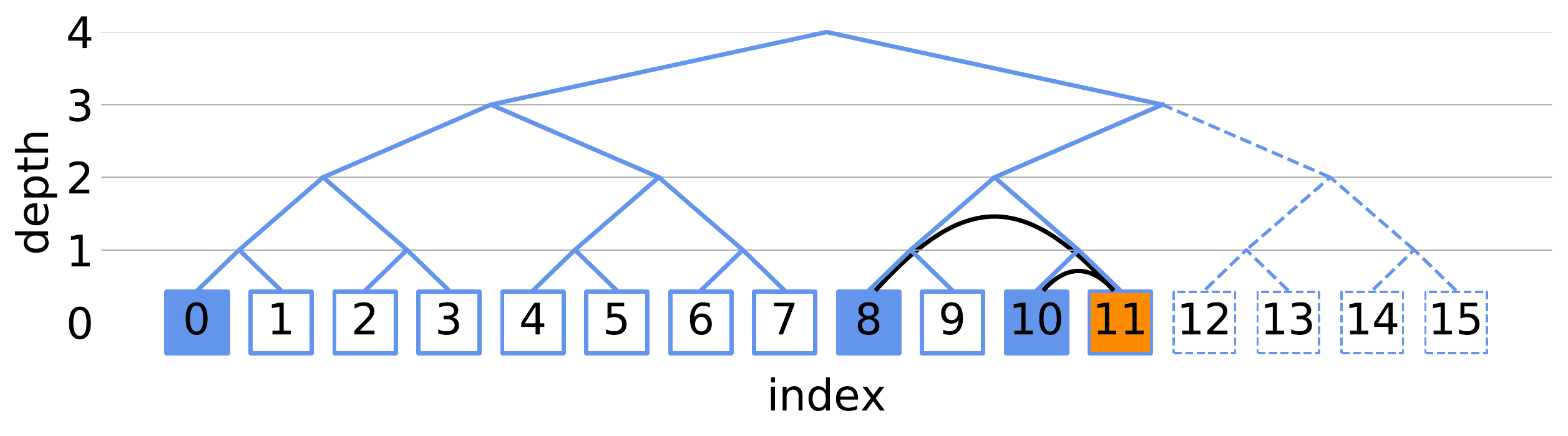}
	\caption{A graphical representation of how binary trees are constructed in \textsc{IterativeBuildTree}. The orange node is the leaf generated at the current step. Blue nodes are the leaves stored in memory for the purpose of checking the U-Turn condition. White nodes are past leaves that have been removed from memory. Dashed white nodes have not been generated yet. Thick black lines link the left and right leaves of subtrees where we need to check the U-Turn condition.}
	\label{fig:binary_tree}
\end{figure}

At an odd step $n$, the data for the candidate nodes indexed by $\mathcal{C}(n)$ must be present in $S$ because the masking procedure ensures that these candidates are the largest even nodes less than $n$ for their corresponding bit counts. Figure \ref{fig:binary_tree} illustrates the iterative procedure at step $n=11$, and full details of the algorithm are provided in Algorithm~\ref{alg:iterative-nuts}.

\DontPrintSemicolon

\begin{figure}
\begin{minipage}[t]{0.49\textwidth}
	\begin{algorithm}[H]
		\caption{\textsc{BuildTree}}
		\label{alg:recursive-nuts}
		{\bf Input} initial node $z$, tree depth $d$ \\
		\eIf{$d = 0$}{
			$z' \from \mathrm{\textsc{Leapfrog}}(z)$ \\
			\Return $\mathrm{\textsc{Tree}}(z', z', False)$ 
		}{
			$T_L \from \mathrm{\textsc{BuildTree}}(z, d - 1)$ \\
			\eIf{$T_L.turning$}{
				\Return $T_L$
			}{
				$z \from T_L.right$ \\
				$T_R \from \mathrm{\textsc{BuildTree}}(z, d - 1)$ \\
				$z_L \from T_L.left$ \\
				$z_R \from T_R.right$ \\
				\eIf{$T_R.turning$}{
					$turning \from True$
				}{
					$turning \from \textrm{\textsc{IsUTurn}}(z_L, z_R)$
				}
				\Return $\mathrm{\textsc{Tree}}(z_L, z_R, turning)$
			}
		}
	\end{algorithm}
\end{minipage}
\hfill
\begin{minipage}[t]{0.47\textwidth}
	\begin{algorithm}[H]
	\caption{\textsc{IterativeBuildTree}}
	\label{alg:iterative-nuts}
		{\bf Input} initial node $z$, tree depth $d$ \\
		{\bf Initialize} storage $S[0], S[1], ..., S[d-1]$ \\
		\For{$n \from 0$ \KwTo $2^d - 1$}{
			$z \from \mathrm{\textsc{Leapfrog}}(z)$ \\
			\eIf{$n$ is even}{
				$ i \from \mathrm{\textsc{BitCount}}(n) $ \\
				$S[i] \from z$
			}{
			    // gets the number of candidate nodes \\
				$l \from \mathrm{\textsc{TrailingBit}}(n)$ \\
                $ i_{max} \from \mathrm{\textsc{BitCount}}(n - 1) $ \\
				$i_{min} \from i_{max} - l + 1$ \\
				\For{$k \from i_{max}$ \KwTo $i_{min}$}{
					$turning \leftarrow \mathrm{\textsc{IsUTurn}}(S[k], z)$  \\
					\If{$turning$}{
						\Return $\mathrm{\textsc{Tree}}(S[0], z, True)$
					}
				}
			}
		}
		\Return $\mathrm{\textsc{Tree}}(S[0], z, False)$
	\end{algorithm}
\end{minipage}
\caption{Comparing the recursive (left) and iterative (right) versions of the tree building algorithm in NUTS. Note that these are high-level specifications of the full algorithms and ignore details about additional metadata in \texttt{Tree} such as the proposal candidate. Other details like the step size and choosing between forward and reverse directions are also omitted, since they are not relevant to the proposed changes.}
\end{figure}

%% file: experiment_detail.tex
\section{Experimental Details}\label{sec:experimental-detail}

All experiments are conducted on a system using an AMD Ryzen Threadripper 1920X processor and an NVIDIA GeForce RTX 2080 Ti graphics card. Framework versions: PyStan $2.19.1.1$, Pyro $1.0$ (with PyTorch $1.3.1$), NumPyro $0.2.3$ (with JAX $0.1.53$ and jaxlib $0.1.36$). For each experiment, we conduct $5$ runs with different random seeds and report the average across $5$ runs. Code used to benchmark all experiments can be found on the \verb|benchmarks| branch of the NumPyro GitHub repository.\footnote{\url{https://github.com/pyro-ppl/numpyro/tree/benchmarks-20191222/benchmarks}}

\paragraph{Hidden Markov Model} To test performance in the small dataset regime, we use an HMM. Following \cite[Section 2.6]{stan_development_team_stan_2018}, we construct a semi-supervised HMM model with 3-dimensional latent states and 10-dimensional observations. Using fixed transition and emission matrices, we sample $600$ data points and treat the first $100$ latent states as observed. For benchmarking we take $1000$ warmup steps and draw $1000$ NUTS samples for both Stan and NumPyro. Because Pyro's NUTS implementation is extremely slow on this problem, we fix the step size to $0.1$ and only draw $40$ samples for each run.

\paragraph{Logistic Regression} 
We consider a logistic regression model on the \verb|Forest CoverType| dataset, which has $581,012$ datapoints and $54$ features.
The prior on the weights is a unit normal distribution.
Following \cite{tran2018simple}, we normalize all features and transform the multi-class problem into a binary class problem by merging all the classes except for the most frequent one. For benchmarking we fix the step size in all frameworks to $0.0015$ and draw $40$ samples. That step size value was obtained by running $150$ warmup adaptation steps in NumPyro.

\paragraph{Sparse Kernel Interaction Model (SKIM)} 
For each value of dimensionality $p$, we produce an artificial dataset with $N=200$ data points that contains 3 randomly selected pairwise interaction terms amongst the $p$ covariates. For each of the frameworks, we adapt the step size and mass matrix using 1000 warmup adaptation steps and then compute the time per effective sample based on the next 1000 drawn samples averaged over 5 random runs.

%% file: vectorized_elbo.tex
\section{Vectorized Estimation of the Evidence Lower Bound (ELBO) in SVI}
\label{sec:vectorized-elbo}

\begin{minipage}{\linewidth}
\begin{lstlisting}[caption={Using \texttt{vmap} to vectorize the computation of the Evidence Lower Bound (ELBO) in SVI.}]
from numpyro import optim
from numpyro.infer import ELBO, SVI

# model, guide defined externally for the inference problem

# vectorize elbo estimate over `num_particles`
class VectorizedELBO:
  def __init__(self, num_particles):
    self.num_particles = num_particles

  # args are arguments passed by SVI to `ELBO().loss`
  def loss(self, rng_key, *args):
    rng_keys = random.split(rng_key, self.num_particles)
    return np.mean(vmap(lambda rng_: ELBO().loss(rng_, *args))(rng_keys))

optimizer = optim.Adam(1e-3)
svi = SVI(model, guide, optimizer, VectorizedELBO(num_particles=100))

\end{lstlisting}
\end{minipage}